\documentclass{article}

\usepackage[utf8]{inputenc} 
\usepackage[T1]{fontenc}    
\usepackage{hyperref}       
\usepackage{url}            
\usepackage{booktabs}       
\usepackage{amsfonts}       
\usepackage{nicefrac}       
\usepackage{microtype}      
\usepackage{times}
\usepackage{epsfig}
\usepackage{graphicx}
\usepackage{amsmath}
\usepackage{amssymb}
\usepackage{algorithmic}
\usepackage[linesnumbered,ruled,vlined]{algorithm2e}
\usepackage{multirow}
\usepackage{booktabs}
\usepackage{siunitx}
\usepackage{bm}
\usepackage{amsfonts}
\usepackage{color}

\title{Robust Data Geometric Structure Aligned Close yet Discriminative Domain Adaptation}

\author{Lingkun Luo$^{1\ast}$,   Xiaofang Wang $^{2}$ ,   Shiqiang Hu $^{1}$, Liming Chen $^{2}$  \\
	$^1 $ School of Aeronautics and Astronautics,\\ Shanghai Jiao Tong University, Shanghai, China. \\ $\{lolinkun1988, sqhu\}$@sjtu.edu.cn \\
	$^2$  LIRIS, CNRS UMR 5205, \'Ecole Centrale de Lyon,\\ 36 avenue Guy de Collongue, \'Ecully, F-69134, France. \\ $\{xiaofang.wang, liming.chen\}$@ec-lyon.fr
}

\begin{document}

\maketitle

\begin{abstract}
Domain adaptation (DA) is transfer learning which aims to leverage labeled data in a related source domain to achieve informed knowledge transfer and help the classification of unlabeled data in a target domain. In this paper, we propose a novel DA method, namely Robust Data Geometric Structure Aligned, Close yet Discriminative Domain Adaptation (RSA-CDDA), which brings closer, in a latent joint subspace, both source and target data distributions, and aligns inherent hidden source and target data geometric structures while performing discriminative DA in repulsing both interclass source and target data. The proposed method performs domain adaptation between source and target in solving a unified model, which incorporates data distribution constraints, in particular via a nonparametric distance, \textit{i.e.}, Maximum Mean Discrepancy (MMD), as well as constraints on inherent hidden data geometric structure segmentation and alignment between source and target, through low rank and sparse representation. RSA-CDDA achieves the search of a joint subspace in solving the proposed unified model through iterative optimization, alternating Rayleigh quotient algorithm and inexact augmented Lagrange multiplier algorithm. Extensive experiments carried out on standard DA benchmarks, \textit{i.e.},  16 cross-domain image classification tasks, verify the effectiveness of the proposed method, which consistently outperforms the state-of-the-art methods.

\end{abstract}

\section{Introduction}

Supervised machine learning requires large amount of labeled training data for an effective training, especially when the underlying prediction model is complex, \textit{e.g.}, deep learning models\cite{long2015learning,glorot2011domain} with a number of parameters at a scale of millions. Furthermore, it assumes that training and testing data have a same distribution for the effectiveness of the learned prediction model. However, such a requirement is hardly satisfied in real-life applications, as labeled data generally are rare and manual annotation could be tedious\cite{7078994,pan2010survey}, expensive and even imprecise or impractical, especially when labeled data require pixel-wise precision in a number of computer vision tasks, \textit{e.g.}, object edge detection, semantic segmentation or medical image segmentation. Transfer learning (TL) aims to mitigate or even bypass such labeled data starvation in leveraging existing related labeled source data. As such, TL has received an increasing interest from various research communities\cite{pan2010survey}. In this paper, we investigate a specific TL problem, namely unsupervised Domain Adaptation (DA), which assumes a shared task space,  a source domain with labeled data and a target domain only with unlabeled data. While the source and target domains are related, their data distributions are assumed to be very different. As a result, a learned prediction model using the labeled source data generally leads to a very poor performance when it is directly applied to the target unlabeled data. Because DA is certainly one of the most frequently   encountered TL problem in real-life applications, it has been the focus of significant research efforts in recent years\cite{glorot2011domain,long2015learning,DBLP:journals/corr/LuoWHWTC17,7078994}.           
  
\begin{figure}[h]
  \centering
\includegraphics[width=1\linewidth]{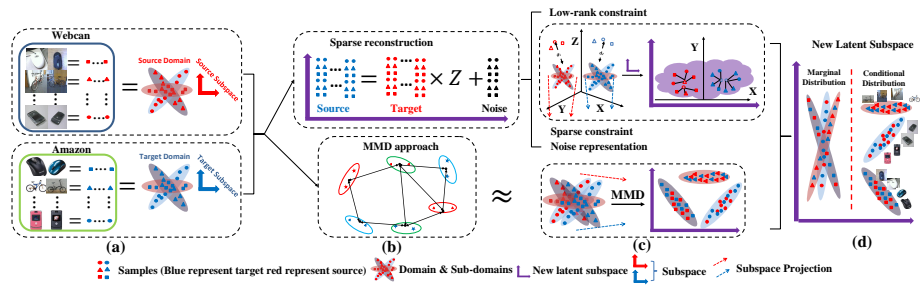}
  \caption{Illustration of the proposed RSA-CDDA method. Fig.1(a): source and target data, \textit{e.g.},   mouse, bike, smartphone images, with different distributions and inherent hidden data geometric structures; Fig.1(b-c) top: low rank and sparse reconstruction of target data by source data to explicit and align data geometric structures between source and target; Fig.1(b-c) bottom: closering data distributions while repulsing interclass source and target data via the nonparametric distance, \textit{i.e.}, Maximum Mean Discrepancy (MMD); Fig.1(d): the achieved latent joint subspace where both marginal and class conditional data distributions are aligned between source and target as well as their data geometric structures; Furthermore, data instances of different classes are well isolated each other, thereby enabling discriminative domain adaptation.      
}

\end{figure}  
  
The mainstream research in DA is the search of a latent joint subspace\cite{pan2011domain} between source and target and features two main lines of approaches. The first line of approaches, \textit{e.g.}, LTSL\cite{DBLP:journals/ijcv/ShaoKF14} , LRSR \cite{DBLP:journals/tip/XuFWLZ16}, seeks a subspace where source and target data can be well aligned and interlaced in preserving inherent hidden geometric data structure via low rank constraint and/or sparse representation, whereas the second line of research, \textit{e.g.}, JDA \cite{long2013transfer}, CDDA \cite{DBLP:journals/corr/LuoWHWTC17}, searches a subspace where the discrepancy between the source and target data distributions is minimized via a nonparametric distance, \textit{i.e.}, Maximum Mean Discrepancy (MMD)\cite{DBLP:conf/nips/GrettonBRSS06}. While the first line of approaches has no guarantee that data distribution discrepancy between the source and target will be minimized, the second one needs to resort to additional ad-hoc methods to further explicit hidden data geometric structure and enable effective label propagation.

In this paper we propose a novel DA method, namely Robust Data Structure Aligned Close yet Discriminative DA (RSA-CDDA), which unifies in a single model the previous two lines of approaches. Specifically, the proposed DA method incorporates into its model objective functions, which search a joint subspace, 1) to bring closer, via a nonparametric distance, \textit{i.e.}, Maximum Mean Discrepancy (MMD), both  marginal and class conditional data distributions while repulsing interclass both source and target data; 2) to achieve alignment of inherent hidden data geometric structures between source and target through locality aware, \textit{i.e.}, low rank,  and sparse reconstruction of the target data using source data; 3) to provide robustness in modeling data outliers through a column-wise reconstruction error matrix which is enforced to be sparse. The resolution of the proposed model is achieved through iterative optimization, alternating Rayleigh quotient algorithm and inexact augmented Lagrange multiplier algorithm. Extensive experiments carried out on 16 cross-domain image classification tasks verify the effectiveness of the proposed DA method, which consistently outperforms state of the art methods. Figure 1 intuitively illustrates the proposed method.      

To sum up, the contributions of this paper are threefold: (1)  design of a novel discriminative DA model, which unifies in a single framework alignment of both data distributions and geometric structures between source and target while repulsing interclass data; (2) introduction of a novel DA algorithm, which searches iteratively a joint subspace optimizing the proposed DA model, using alternatively Rayleigh quotient algorithm and inexact augmented Lagrange multiplier algorithm; (3) Verification of the effectiveness of the proposed DA method through 16 cross-domain image classification tasks.

\section{Related Work}


State of the art in DA has featured so far two main approaches: 1) the search of a novel joint subspace where source and target domain have a similar data distribution \cite{pan2011domain,long2013transfer,DBLP:journals/corr/LuoWHWTC17}; or 2) adaptation of the prediction model trained using the source data \cite{DBLP:journals/tip/XuFWLZ16,DBLP:journals/ijcv/ShaoKF14,DBLP:conf/cvpr/JhuoLLC12}. In the first approach, source and target data are changed because projected into the novel joint subspace. A prediction model trained on labeled source data in this novel subspace can thus be applied to projected target data because they are aligned or have similar data distribution in the novel joint subspace.  The second approach  is to modify the parameters of the prediction model trained on the source domain so that the decision boundaries are adapted to the target data which remain unchanged.   

Because target domain only contains unlabeled data, modification of a prediction model's parameters proves difficult and an increasing research focus has recently turned on the first approach. One can distinguish two lines of research work in this direction: 1) Data distribution oriented methods \cite{pan2008transfer,pan2011domain,long2013transfer,DBLP:journals/corr/LuoWHWTC17},  ; and 2) Data reconstruction-based methods \cite{DBLP:journals/tip/XuFWLZ16,DBLP:journals/ijcv/ShaoKF14,DBLP:conf/cvpr/JhuoLLC12,DBLP:journals/tip/ZhangZZ16}.

Data distribution oriented methods seek a joint subspace to decrease the mismatch between the source and target data distributions, using the nonparametric metric, namely Maximum Mean Discrepancy (MMD), based on Reproducing Hilbert Space (RKHS)\cite{DBLP:conf/ismb/BorgwardtGRKSS06}. TCA \cite{pan2011domain}  \cite{pan2008transfer}  brings closer the source and target marginal distributions; JDA \cite{long2013transfer} further decreases  the discrepancy of both the marginal and conditional distributions. CDDA \cite{DBLP:journals/corr/LuoWHWTC17} goes one step further with respect to JDA and achieves discriminative DA in introducing a repulsive force term in their model to repulse interclass instances. However, all these methods do not account explicitly in their model for the inherent hidden data structure which are important for reliable label propagation,  thereby avoiding negative knowledge transfer from the source domain. One exception is CDDA, which, however, resorts to an ad-hoc method,  namely spectral clustering, to make the label propagation respect a constraint of Geometric Structure Consistency. 

This drawback has been precisely addressed by data reconstruction-based methods, \textit{e.g.}, RDALR\cite{DBLP:conf/cvpr/JhuoLLC12}, LTSL\cite{DBLP:journals/ijcv/ShaoKF14}, LRSR\cite{DBLP:journals/tip/XuFWLZ16} and LSDT\cite{DBLP:journals/tip/ZhangZZ16},  which search a learned subspace in minimizing the reconstruction error of target data using source or both source and target data, and  make use of low rank and sparse representation to segment the inherent hidden data structure and account for data outliers as well. While these methods ensure that source and target data are well aligned and interleaved, they do not have theoretic guarantee that aligned source and target data have similar data distribution.

The proposed RSA-CDDA combines the advantages of the previous two research lines  and unifies in a same framework both the data distribution oriented methods and reconstruction-based ones.

\section{Method}

\subsection{Notations and Problem Statement}

We begin with the definitions of notations and concepts most of which we borrow directly  from CDDA \cite{DBLP:journals/corr/LuoWHWTC17}. 

Vectors and matrices are frequently used in the subsequent and represented using bold symbols. Given a matrix $\bf{M}$, we define the Frobenius norm ${\left\| . \right\|_F}$ and nuclear norm ${\left\| . \right\|_*}$ as: ${\left\| {\bf{M}} \right\|_F} = {\left\| {\sigma ({\bf{M}})} \right\|_2}$, ${\left\| {\bf{M}} \right\|_*} = {\left\| {\sigma ({\bf{M}})} \right\|_1}$, where ${\sigma ({\bf{M}})}$  is the singular values vector of the matrix $\bf{M}$.

A domain $D$ is defined as an m-dimensional feature space $\chi$ and a marginal probability distribution $P(x)$, \textit{i.e.}, $\mathcal{D}=\{\chi,P(x)\}$ with $x\in \chi$. 

Given a specific domain $D$, a  task $T$ is composed of a C-cardinality label set $\mathcal{Y}$  and a classifier $f(x)$,\textit{ i.e.}, $T = \{\mathcal{Y},f(x)\}$, where $f({x}) = \mathcal{Q}( y |x)$ which can be interpreted as the class conditional probability distribution for each input sample $x$. 
%
%

In unsupervised domain adaptation (DA), we are given a source domain $\mathcal{D_S}=\{x_{i}^{s},y_{i}^{s}\}_{i=1}^{n_s}$ with $n_s$ labeled samples, and  a unlabeled target domain $\mathcal{D_T}=\{x_{j}^{t}\}_{j=1}^{n_t}$ with $n_t$  unlabeled samples with the assumption that source domain $\mathcal{D_S}$ and target domain $\mathcal{D_T}$ are different, \textit{i.e.},  $\mathcal{\chi}_S=\mathcal{{\chi}_T}$, $\mathcal{Y_S}=\mathcal{Y_T}$, $\mathcal{P}(\mathcal{\chi_S}) \neq \mathcal{P}(\mathcal{\chi_T})$, $\mathcal{Q}(\mathcal{Y_S}|\mathcal{\chi_{S}}) \neq \mathcal{Q}(\mathcal{Y_T}|\mathcal{\chi_{T}})$.
We also define the notion of sub-domain, denoted as ${\cal D}_{\cal S}^{(c)}$, representing the set of samples in ${{\cal D}_{\cal S}}$ with label $c$. Similarly, a sub-domain ${\cal D}_{\cal T}^{(c)}$ can be defined for the target domain as the set of samples in ${{\cal D}_{\cal T}}$ with label $c$. However, as ${{\cal D}_{\cal T}}$is the target domain with unlabeled samples, a basic classifier,\textit{ e.g.}, Nearest Neighbor (NN), is needed to attribute  pseudo labels for samples in ${{\cal D}_{\cal T}}$.


The aim of the Robust Data Geometric Structure Aligned Close yet Discriminative Domain Adaptation (RSA-CDDA) is to learn a latent subspace  with the following properties: P1) the discrepancy of both the marginal and conditional distributions between the source and target domains is reduced; P2)  The distances between each sub-domain to the others,  are increased in order to  enable a discriminative DA; P3) both the inherent local and global data geometric structures are preserved and aligned for reliable label prediction; and P4) Data outliers are accounted for to avoid negative transfer.

\subsection{Model}
\subsubsection{Latent Feature Space with Dimensionality Reduction}
The finding of a latent feature space  with dimensionality reduction  has been demonstrated useful for DA in several previous works, \textit{e.g.}, \cite{pan2008transfer,pan2011domain,long2013transfer}.  One of its important properties is that  original data is projected into a lower dimensional space which is considered as \emph{principal} structure of data. In the proposed method, we also apply the Principal Component Analysis (PCA).  Mathematically, given with an input data matrix $\boldsymbol{X} = [{\mathcal{D_S}},\mathcal{D_T}]$, $\boldsymbol{X} \in {\mathbb{R}^{m\times({n_s} + {n_t})}}$, the centering matrix is defined as  $\boldsymbol{H} = \boldsymbol{I} - \frac{1}{n_s+n_t}\boldsymbol{1}$, where $\boldsymbol{1}$ is the $(n_s+n_t) \times (n_s+n_t)$ matrix of ones. The optimization of PCA is to find a projection space $\boldsymbol{A}$ which  maximizes the embedded data variance.
\begin{equation}\label{eq:pca}
	\begin{array}{c}
		\mathop {\max}\limits_{\boldsymbol{A^TA} = \boldsymbol{I}} tr(\boldsymbol{A}^T\boldsymbol{ XH}\boldsymbol{X}^T \boldsymbol{A})
	\end{array}
\end{equation}
where $tr(\mathord{\cdot})$ denotes the trace of a matrix,   $\boldsymbol{XH}\boldsymbol{X}^T$ is the data covariance matrix, and $\bf A \in \mathbb{R}^{m \times k}$ with $m$ the feature dimension and $k$ the dimension of the projected subspace. The optimal solution  is calculated by solving an eigendecomposition problem: $\boldsymbol{XH}\boldsymbol{X}^T=\boldsymbol{A\Phi}$, where $\boldsymbol{\Phi}=diag(\phi_1,\dots, \phi_k )$ are the $k$ largest eigenvalues. Finally, the original data $\boldsymbol{X}$ is projected into the  optimal $k$-dimensional subspace using $\boldsymbol{Z} = \boldsymbol{A}^T\boldsymbol{X}$. 

\subsubsection{ Closering Marginal and Conditional Distributions}

However, the subspace calculated via PCA does not decrease the mismatch of data distributions between the source and target domain. As a result, to meet property P1, we explicitly leverage the nonparametric distance measurement MMD in RKHS \cite{borgwardt2006integrating} to compute the distance between expectations of source domain/sub-domain and target domain/sub-domain, once the original data projected into  a low-dimensional feature space.  Formally, the empirical distance of the source and target domains are defined as $Dis{t^{marginal}}$. The distance of conditional probability distributions $Dis{t^{conditional}}$ is defined as the sum of the empirical distances over the class labels between the sub-domains of a same label in the source and target domain:
 \begin{equation}\label{eq:JDA}
 	\resizebox{0.95\hsize}{!}{%
 		$\begin{array}{*{20}{l}}
 		{Dis{t_{Close}} = Dis{t^{marginal}}({D_S},{D_T}) + Dis{t^{conditional}}\sum\limits_{c = 1}^C {({D_S}^c,{D_T}^c)}  = }\\
 		{{{\left\| {\frac{1}{{{n_s}}}\sum\limits_{i = 1}^{{n_s}} {{{\bf{A}}^T}{x_i} - } \frac{1}{{{n_t}}}\sum\limits_{j = {n_s} + 1}^{{n_s} + {n_t}} {{{\bf{A}}^T}{x_j}} } \right\|}^2} + {{\left\| {\frac{1}{{n_s^{(c)}}}\sum\limits_{{x_i} \in {D_S}^{(c)}} {{{\bf{A}}^T}{x_i}}  - \frac{1}{{n_t^{(c)}}}\sum\limits_{{x_j} \in {D_T}^{(c)}} {{{\bf{A}}^T}{x_j}} } \right\|}^2} = tr({{\bf{A}}^T}{\bf{X}}({{\bf{M}}_{\bf{0}}} + \sum\limits_{c = 1}^{c = C} {{{\bf{M}}_c}} ){{\bf{X}}^{\bf{T}}}{\bf{A}})}
 		\end{array}$}
 \end{equation}
where $C$ is the number of classes, $\mathcal{D_S}^{(c)} = \{ {x_i}:{x_i} \in \mathcal{D_S} \wedge y({x_i} = c)\} $ represents the ${c^{th}}$ sub-domain in the source domain, $n_s^{(c)} = {\left\| {\mathcal{D_S}^{(c)}} \right\|_0}$ is the number of samples in the ${c^{th}}$ {source} sub-domain. $\mathcal{D_T}^{(c)}$ and $n_t^{(c)}$ are defined similarly for the target domain. Finally, ${{\bf{M}}_0}$ represents the marginal distribution between ${{\cal D}_{\cal S}}$ and ${{\cal D}_{\cal T}}$ and $\bf M_c$ represents the conditional distribution between sub-domains in ${{\cal D}_{\cal S}}$ and ${{\cal D}_{\cal T}}$, they are defined as:
 \begin{equation}\label{eq:MC}
 	\resizebox{1\hsize}{!}{%
 		$\begin{array}{*{20}{l}}
 		{{{({{\bf{M}}_0})}_{ij}} = \left\{ {\begin{array}{*{20}{l}}
 				{\frac{1}{{{n_s}{n_s}}},\;\;\;{x_i},{x_j} \in {D_S}}\\
 				{\frac{1}{{{n_t}{n_t}}},\;\;\;{x_i},{x_j} \in {D_T}}\\
 				{0,\;\;\;\;\;\;\;\;\;\;\;\;otherwise}
 				\end{array}} \right.}
 		\end{array}\;;\;\;{({{\bf{M}}_c})_{ij}} = \left\{ {\begin{array}{*{20}{l}}
 			{\frac{1}{{n_s^{(c)}n_s^{(c)}}},\;\;\;{x_i},{x_j} \in {D_S}^{(c)}}\\
 			{\frac{1}{{n_t^{(c)}n_t^{(c)}}},\;\;\;{x_i},{x_j} \in {D_T}^{(c)}}\\
 			{\frac{{ - 1}}{{n_s^{(c)}n_t^{(c)}}},\;\;\;\left\{ {\begin{array}{*{20}{l}}
 					{{x_i} \in {D_S}^{(c)},{x_j} \in {D_T}^{(c)}}\\
 					{{x_i} \in {D_T}^{(c)},{x_j} \in {D_S}^{(c)}}
 					\end{array}} \right.}\\
 			{0,\;\;\;\;\;\;\;\;\;\;\;\;otherwise}
 			\end{array}} \right.$}
 \end{equation}
The difference between the marginal distributions $\mathcal{P}(\mathcal{X_S})$ and $\mathcal{P}(\mathcal{X_T})$ is reduced in minimizing {$Dis{t^{marginal}}({{\cal D}_{\cal S}},{{\cal D}_{\cal T}})$} and the mismatch of conditional distributions between ${{D_{\cal S}}^c}$ and ${{D_{\cal T}}^c}$ is reduced in minimizing ${Dis{t^{conditional}}\sum\limits_{c = 1}^C {({D_{\cal S}}^c,{D_{\cal T}}^c)} }$.

\subsubsection{Repulsing interclass data for discriminative DA}

In bringing closer data distributions, the previous DA model does not explicitly achieve a discriminative DA in repulsing interclass data. Such a discriminative DA can be achieved by introducing a novel \textit{repulsive force} , which aims to increase interclass distances, thereby satisfying property P2. Specifically, the repulsive force for DA is defined as: 
$Dis{t^{repulsive}} = Dist_{{\cal S} \to {\cal T}}^{repulsive} + Dist_{{\cal T} \to {\cal S}}^{repulsive} + Dist_{{\cal S} \to {\cal S}}^{repulsive}$, where ${{\cal S} \to {\cal T}}$ , ${{\cal T} \to {\cal S}}$ and ${{\cal S} \to {\cal S}}$ index the distances computed from ${D_{\cal S}}$ to ${D_{\cal T}}$ , ${D_{\cal T}}$ to ${D_{\cal S}}$ and ${D_{\cal S}}$ to ${D_{\cal S}}$ respectively. $Dist_{{\cal S} \to {\cal T}}^{repulsive}$ represents the sum of the distances between each source sub-domain ${D_{\cal S}}^{(c)}$ and all the  target sub-domains ${D_{\cal T}}^{(r);\;r \in \{ \{ 1...C\}  - \{ c\} \} }$ except the one with the label $c$. $Dist_{{\cal T} \to {\cal S}}^{repulsive}$ represents the sum of the distances from each target sub-domain ${D_{\cal T}}^{(c)}$ to all the the source sub-domains ${D_{\cal S}}^{(r);\;r \in \{ \{ 1...C\}  - \{ c\} \} }$ except the source sub-domain with the label $c$. $Dist_{{\cal S} \to {\cal S}}^{repulsive}$ represents the sum of the distances from each source sub-domain ${D_{\cal S}}^{(c)}$ to all the the source sub-domains ${D_{\cal S}}^{(r);\;r \in \{ \{ 1...C\}  - \{ c\} \} }$ except the source sub-domain with the label $c$. The sum of these distances is explicitly defined as:
	\begin{equation}\label{eq:CDDAnew}
		\resizebox{1.02\hsize}{!}{%
			$\begin{array}{l}
			\begin{array}{*{20}{l}}
			{Dis{t^{repulsive}} = Dist_{S \to T}^{repulsive} + Dist_{T \to S}^{repulsive} + Dist_{S \to S}^{repulsive} = \sum\limits_{c = 1}^C {tr({{\bf{A}}^T}{\bf{X}}({{\bf{M}}_{S \to T}} + {{\bf{M}}_{T \to S}} + {{\bf{M}}_{S \to S}}){{\bf{X}}^{\bf{T}}}{\bf{A}})} }\\
			{\sum\limits_{c = 1}^C {\begin{array}{*{20}{l}}
					{{{\left\| {\frac{1}{{n_s^{(c)}}}\sum\limits_{{x_i} \in {D_S}^{(c)}} {{{\bf{A}}^T}{x_i}}  - \frac{1}{{\sum\limits_{r \in \{ \{ 1...C\}  - \{ c\} \} } {n_t^{(r)}} }}\sum\limits_{{x_j} \in D_T^{(r)}} {{{\bf{A}}^T}{x_j}} } \right\|}^2} + \sum\limits_{c = 1}^C {\begin{array}{*{20}{l}}
							{{{\left\| {\frac{1}{{n_s^{(c)}}}\sum\limits_{{x_i} \in {D_T}^{(c)}} {{{\bf{A}}^T}{x_i}}  - \frac{1}{{\sum\limits_{r \in \{ \{ 1...C\}  - \{ c\} \} } {n_t^{(r)}} }}\sum\limits_{{x_j} \in D_S^{(r)}} {{{\bf{A}}^T}{x_j}} } \right\|}^2}}
							\end{array}} }
					\end{array}} }
			\end{array}\\
			{\rm{ + }}\sum\limits_{c = 1}^C {\begin{array}{*{20}{l}}
				{{{\left\| {\frac{1}{{n_s^{(c)}}}\sum\limits_{{x_i} \in {D_S}^{(c)}} {{{\bf{A}}^T}{x_i}}  - \frac{1}{{\sum\limits_{r \in \{ \{ 1...C\}  - \{ c\} \} } {n_t^{(r)}} }}\sum\limits_{{x_j} \in D_S^{(r)}} {{{\bf{A}}^T}{x_j}} } \right\|}^2}}
				\end{array}} 
			\end{array}$}
	\end{equation}
		   where ${{\bf{M}}_{{\cal S} \to {\cal T}}}$ , ${{\bf{M}}_{{\cal T} \to {\cal S}}}$ and ${{\bf{M}}_{{\cal S} \to {\cal S}}}$ are defined as
		\begin{equation}\label{eq:Mmatrix}
		\resizebox{1\hsize}{!}{%
			$\begin{array}{l}
			{({{\bf{M}}_{{\bf{S}} \to {\bf{T}}}})_{ij}} = \left\{ {\begin{array}{*{20}{l}}
				{\frac{1}{{n_s^{(c)}n_s^{(c)}}},\;\;\;{x_i},{x_j} \in {D_S}^{(c)}}\\
				{\frac{1}{{n_t^{(r)}n_t^{(r)}}},\;\;\;{x_i},{x_j} \in {D_T}^{(r)}}\\
				{\frac{{ - 1}}{{n_s^{(c)}n_t^{(r)}}},\;\;\;\left\{ {\begin{array}{*{20}{l}}
						{{x_i} \in {D_S}^{(c)},{x_j} \in {D_T}^{(r)}}\\
						{{x_i} \in {D_T}^{(r)},{x_j} \in {D_S}^{(c)}}
						\end{array}} \right.}\\
				{0,\;\;\;\;\;\;\;\;\;\;\;\;otherwise}
				\end{array}} \right.;
                \;\;\;{({{\bf{M}}_{{\bf{T}} \to {\bf{S}}}})_{ij}} = \left\{ {\begin{array}{*{20}{l}}
				{\frac{1}{{n_t^{(c)}n_t^{(c)}}},\;\;\;{x_i},{x_j} \in {D_T}^{(c)}}\\
				{\frac{1}{{n_s^{(r)}n_s^{(r)}}},\;\;\;{x_i},{x_j} \in {D_S}^{(r)}}\\
				{\frac{{ - 1}}{{n_t^{(c)}n_s^{(r)}}},\;\;\;\left\{ {\begin{array}{*{20}{l}}
						{{x_i} \in {D_T}^{(c)},{x_j} \in {D_S}^{(r)}}\\
						{{x_i} \in {D_S}^{(r)},{x_j} \in {D_T}^{(c)}}
						\end{array}} \right.;}\\
				{0,\;\;\;\;\;\;\;\;\;\;\;\;otherwise}
				\end{array}} \right.
			{({{\bf{M}}_{{\bf{S}} \to {\bf{S}}}})_{ij}} = \left\{ {\begin{array}{*{20}{l}}
				{\frac{1}{{n_s^{(c)}n_s^{(c)}}},\;\;\;{x_i},{x_j} \in {D_S}^{(c)}}\\
				{\frac{1}{{n_s^{(r)}n_s^{(r)}}},\;\;\;{x_i},{x_j} \in {D_S}^{(r)}}\\
				{\frac{{ - 1}}{{n_s^{(c)}n_s^{(r)}}},\;\;\;\left\{ {\begin{array}{*{20}{l}}
						{{x_i} \in {D_S}^{(c)},{x_j} \in {D_S}^{(r)}}\\
						{{x_i} \in {D_S}^{(r)},{x_j} \in {D_S}^{(c)}}
						\end{array}} \right.}\\
				{0,\;\;\;\;\;\;\;\;\;\;\;\;otherwise}
				\end{array}} \right.
			\end{array}$}
		\end{equation}

  Finally, we obtain
	\begin{equation}\label{eq:repulsive}
		\resizebox{0.65\hsize}{!}{%
			 ${Dist}^{repulsive} = \sum\limits_{c = 1}^C {tr({{\bf{A}}^T}{\bf{X}}({{\bf{M}}_{S \to T}} + {{\bf{M}}_{T \to S}} + {{\bf{M}}_{S \to S}}){{\bf{X}}^{\bf{T}}}{\bf{A}})} $}
	\end{equation}
We define ${{\bf{M}}_{REP}} = {{\bf{M}}_{S \to T}} + {{\bf{M}}_{T \to S}} + {{\bf{M}}_{S \to S}} $ as the \textit{repulsive force} constraint matrix. While the minimization of Eq.(\ref{eq:JDA})  makes closer both marginal and conditional distributions between source and target, the maximization of Eq.(\ref{eq:CDDAnew}) increases the distances between source and target sub-domains with different labels as well as source sub-domains with different labels, thereby enhancing the discriminative power of the underlying latent feature space.

\subsubsection{Data geometric structure alignment}
Both source and target data can be embedded into a manifold with complex geometric structure. While the model developed in the previous subsections brings closer data distributions while repulsing interclass data, it does not explicitly account for the inherent hidden data geometric structure. We tackle this problem so that our DA model enables data geometric structure alignment between source and target and thereby meets property P3. For this purpose, we propose to use source data 	${{\bf{X}}_s}$ to linearly reconstruct target data ${{\bf{X}}_t}$ in a common latent subspace in learning a reconstruction coefficient matrix $\textbf{Z}$. In noting $A$ the projection transformation, the reconstruction problem can be formulated as:  ${{\bf{A}}^T}{{\bf{X}}_t} = {{\bf{A}}^T}{{\bf{X}}_s}{\bf{Z}}$. 

To further align data geometric structures between source and target, we introduce into our model two additional constraints, namely locality aware and sparse representation constraints. The locality constraint is a constraint already widely explored in manifold learning \cite{7078994,DBLP:journals/ijcv/ShaoKF14,DBLP:journals/tip/XuFWLZ16,DBLP:journals/corr/LuoWHWTC17}. It aims to ensure that target data in a neighborhood is only reconstructed from neighboring source data and thereby intuitively preserves and aligns source and target data geometric structures. This locality constraint is achieved by enforcing the reconstruction matrix $\textbf{Z}$ to be low rank with a block-wise structure. As a result, the reconstruction problem aware of locality is now defined as:$\mathop {\min }\limits_{{\bf{A}},{\bf{Z}}} rank({\bf{Z}})\;\;s.t.\;{{\bf{A}}^T}{{\bf{X}}_t} = {{\bf{A}}^T}{{\bf{X}}_s}{\bf{Z}}$. However, rank minimization problem is non-convex, which is difficult to solve. Fortunately, \cite{DBLP:journals/corr/LinCM10} points out we could treat the rank constraint problem as nuclear norm problem, and reformulate it as: $\mathop {\min }\limits_{{\bf{A}},{\bf{Z}}} {\left\| {\bf{Z}} \right\|_*}\;\;s.t.\;{{\bf{A}}^T}{{\bf{X}}_t} = {{\bf{A}}^T}{{\bf{X}}_s}{\bf{Z}}$.

The sparse representation constraint aims to further ensure the alignment of data geometric structures between source and target in enforcing that each target datum is only sparsely reconstructed from a few meaningful source data, and thereby source and target data are locally interleaved in the searched subspace. Therefore, the reconstruction problem can be further formulated as : $\mathop {\min }\limits_{{\bf{A}},{\bf{Z}}} {\left\| {\bf{Z}} \right\|_*} + {\left\| {\bf{Z}} \right\|_1}\;s.t.\;{{\bf{A}}^T}{{\bf{X}}_t} = {{\bf{A}}^T}{{\bf{X}}_s}{\bf{Z}} + {\bf{E}}$, with $\bf{E}$ the column-wise error matrix.  

To account for a few data outliers and meet property P4, we can simply enforce the column-wise error matrix $\bf{E}$ to be sparse and thereby provide robustness of the proposed DA method to noisy data and alleviate the influence of negative transfer. As a result, the objective function of our DA method for the alignment of data geometric structures between source and target and robustness to data outliers is defined as follows:
        	\begin{equation}\label{eq:min}
      		\begin{array}{*{20}{c}}
      			{\mathop {\min }\limits_{{\bf{A}},{\bf{Z}},{\bf{E}}} {\lambda _1}{{\left\| {\bf{E}} \right\|}_1} + {{\left\| {\bf{Z}} \right\|}_*} + {\lambda _2}{{\left\| {\bf{Z}} \right\|}_1}\;\;\;\;\;s.t.\;{{\bf{A}}^T}{{\bf{X}}_t} = {{\bf{A}}^T}{{\bf{X}}_s}{\bf{Z}} + {\bf{E}}}
      		\end{array}
      	\end{equation}	

\subsubsection{Final energy function}

In integrating all the properties expressed in the previous subsections, \textit{i.e.},  Eq.(\ref{eq:JDA}), Eq.(\ref{eq:CDDAnew}) and Eq.(\ref{eq:min}), we obtain our designed final DA model, formulated as Eq.(\ref{eq:opt}):  
				  		\vspace{-0.5pt}
	  	\begin{equation}\label{eq:opt}
	  		\resizebox{0.93\hsize}{!}{%
	  			$\begin{array}{*{20}{l}}
	  			{\mathop {\min }\limits_{{\bf{A}},{\bf{E}},{\bf{M}},{\bf{Z}},{{\bf{Z}}_{\bf{l}}},{{\bf{Z}}_{\bf{s}}}} tr({{\bf{A}}^T}{\bf{X}}({{\bf{M}}_{\bf{C}}} - {{\bf{M}}_{{\bf{REP}}}}){{\bf{X}}^T}{\bf{A}}) + \lambda \left\| {\bf{A}} \right\|_F^2 + {\lambda _1}{{\left\| {\bf{E}} \right\|}_{\bf{1}}} + {\lambda _2}{{\left\| {{{\bf{Z}}_{\bf{s}}}} \right\|}_{\bf{1}}} + {{\left\| {{{\bf{Z}}_{\bf{l}}}} \right\|}_*}}\\
	  			{s.t.\;{{\bf{A}}^T}{{\bf{X}}_{\bf{t}}} = {{\bf{A}}^T}{{\bf{X}}_{\bf{s}}}{\bf{Z}} + {\bf{E}},\;{{\bf{Z}}_{\bf{l}}}{\rm{ = }}{\bf{Z}},\;\;{{\bf{Z}}_{\bf{s}}}{\rm{ = }}{\bf{Z}},\;{{\bf{A}}^T}{\bf{XH}}{{\bf{X}}^T}{\bf{A}} = {\bf{I}},\;{{\bf{M}}_{\bf{C}}} = {{\bf{M}}_{\bf{0}}} + \sum\limits_{c = 1}^{c = C} {{{\bf{M}}_c}}  = \sum\limits_{c = 0}^{c = C} {{{\bf{M}}_c}} }
	  			\end{array}$}
	  	\end{equation}
Through iterative optimization of Eq.(\ref{eq:opt}), our DA model searches a latent subspace satisfying properties P1 through P4.

\subsection{Optimization}
We solve Eq.(\ref{eq:opt}) through two main steps. Firstly, Rayleigh quotient algorithm is applied to calculate initial ${{\bf{M}}_\textbf{RSA}} =  {{\bf{M}}_\textbf{C}} - {{\bf{M}}_\textbf{REP}}$. Eq.(\ref{eq:opt}) is iteratively optimized via augmented Lagrange multiplier (ALM) method as in Eq.(\ref{eq:lag}):
	\vspace{-0.5pt}
	\begin{equation}\label{eq:lag}
	\resizebox{0.94\hsize}{!}{%
		$\begin{array}{*{20}{l}}
		{Energy = tr({{\bf{A}}^T}{\bf{X}}({{\bf{M}}_{\bf{RSA}}}){{\bf{X}}^T}{\bf{A}}) + \lambda \left\| {\bf{A}} \right\|_F^2 + {\lambda _1}{{\left\| {\bf{E}} \right\|}_{\bf{1}}} + {\lambda _2}{{\left\| {{{\bf{Z}}_{\bf{s}}}} \right\|}_{\bf{1}}} + {{\left\| {{{\bf{Z}}_{\bf{l}}}} \right\|}_*}}\\
		{ + \left\langle {{{\bf{Y}}_1},{{\bf{A}}^T}{{\bf{X}}_{\bf{t}}} - {{\bf{A}}^T}{{\bf{X}}_s}{\bf{Z}} - {\bf{E}}} \right\rangle  + \left\langle {{{\bf{Y}}_2},{\bf{Z}} - {{\bf{Z}}_{\bf{l}}}} \right\rangle  + \left\langle {{{\bf{Y}}_3},{\bf{Z}} - {{\bf{Z}}_{\bf{s}}}} \right\rangle  + \frac{\mu }{2}\left\| {{{\bf{A}}^T}{{\bf{X}}_{\bf{t}}} - {{\bf{A}}^T}{{\bf{X}}_s}{\bf{Z}} - {\bf{E}}} \right\|_F^2}\\
		{ + \frac{\mu }{2}(\left\| {{\bf{Z}} - {{\bf{Z}}_{\bf{1}}}} \right\|_F^2 + \left\| {{\bf{Z}} - {{\bf{Z}}_2}} \right\|_F^2) + \frac{\mu }{2}tr({{\bf{A}}^T}{\bf{XH}}{{\bf{X}}^T}{\bf{A}} - {\bf{I}})}
		\end{array}$}
	\end{equation}

To solve efficiently the problem defined in Eq.(\ref{eq:opt}), we calculate the initial MMD matrix ${{\bf{M}}_\textbf{RSA}}$ via Rayleigh quotient algorithm\cite{long2013transfer,DBLP:journals/corr/LuoWHWTC17} as shown in Algorithm 1(a). The process to solve the projection matrix $\bf{A}$ is shown in Algorithm 1(b). Step 1 and Step 3 of Algorithm 1(b) are derived through analytical calculation. ${{\bf{M}}_\textbf{RSA}}$ is updated at each iteration of Algorithm 1(b) via the same process as Step 3 in Algorithm 1(a). Step 4, 5 and  6 in Algorithm 1(b) are calculated according to \cite{DBLP:journals/tip/XuFWLZ16,DBLP:journals/pami/LiuLYSYM13,DBLP:journals/corr/LinCM10}. Detail proof procedure are not shown here due to space limitation and will be provided online as supplementary materials.

\begin{algorithm}[!h]
\scriptsize
	\caption{Proposed method}

	\textbf{(a) Initial MMD matrix calculation}
	\rule{\textwidth}{0.3mm}
	
	\KwIn{Data ${\bf{X}} = ({{\bf{X}}_{\bf{t}}} \cup {{\bf{X}}_{\bf{s}}})$, Source domain label ${\bf{Y}}_{\cal S}$, subspace bases $k$, iterations $T$, regularization parameter $\lambda $ and $\alpha $}
	\While{$\sim isempty(\bf{X},{{\bf{Y}}_{\cal S}})$ and $t<T$	}{
		\textbf{Step {1}}:  Construct ${\bf{M}}_{\bf{C}}$ and ${{{\bf{M}}_{\bf{REP}}}}$  ;\\
		\textbf{Step 2}: Projection space calculation \\
		
		(i) Calculate ${{\bf{M}}_{{\bf{RSA}}}} = {{\bf{M}}_{\bf{C}}} - {{{\bf{M}}_{\bf{REP}}}} $;\\
		(ii) Solve the generalized eigendecomposition problem as  $({\bf{X}}{{\bf{M}}_{{\bf{RSA}}}}{{\bf{X}}^T} + \lambda {\bf{I}}){{\bf{A}}_{{\bf{mmd}}}} = {\bf{XH}}{{\bf{X}}^T}{{\bf{A}}_{{\bf{mmd}}}}\Phi  $, and obtain adaptation matrix ${{\bf{A}}_{{\bf{mmd}}}}$;\\
		(iii) Embed data via the transformation, ${{\bf{D}}_{{\bf{mmd}}}} = {\bf{A}}_{{\bf{mmd}}}^{\bf{T}}{\bf{X}}$\;
		
		\textbf{Step 3}: Update ${{\bf{M}}_{{\bf{RSA}}}}$   \\
		(i) train a classifier $f$ and update pseudo target labels ${\bf{Y}}_{\cal T}$\;
	    (ii) update ${\bf{M}}_{\bf{C}}$ and ${{{\bf{M}}_{\bf{REP}}}}$ via Eq.(\ref{eq:Mmatrix}) and Eq.(\ref{eq:MC})\;
		(iii) obtain new ${{\bf{M}}_{{\bf{RSA}}}}$\;
	
	\textbf{Step  4}: Return to Step1;  $t=t+1$;\\               
	
}

\KwOut{MMD matrix ${{\bf{M}}_{{\bf{RSA}}}}$}
	\rule{\textwidth}{0.3mm}
	\textbf{(b) Solve  Optimization Eq.(\ref{eq:lag})  via Inexact ALM}
		\vspace{-2pt}
	\rule{\textwidth}{0.3mm}
	\textbf{Require and Ensure}: $\begin{array}{*{20}{l}}
	{\mathop {\min }\limits_{{\bf{A}},{\bf{E}},{\bf{M}},{\bf{Z}},{{\bf{Z}}_{\bf{l}}},{{\bf{Z}}_{\bf{s}}}} tr({{\bf{A}}^T}{\bf{X}}({{\bf{M}}_{{\bf{RSA}}}}){{\bf{X}}^T}{\bf{A}}) + \lambda \left\| {\bf{A}} \right\|_F^2 + {\lambda _1}{{\left\| {\bf{E}} \right\|}_{\bf{1}}} + {\lambda _2}{{\left\| {{{\bf{Z}}_{\bf{s}}}} \right\|}_{\bf{1}}} + {{\left\| {{{\bf{Z}}_{\bf{l}}}} \right\|}_*}}\\
	{s.t.\;{{\bf{A}}^T}{{\bf{X}}_{\bf{t}}} = {{\bf{A}}^T}{{\bf{X}}_{\bf{s}}}{\bf{Z}} + {\bf{E}},\;\;{{\bf{Z}}_{\bf{l}}}{\rm{ = }}{\bf{Z}},\;\;{{\bf{Z}}_{\bf{s}}}{\rm{ = }}{\bf{Z}},\;\;{{\bf{A}}^T}{\bf{XH}}{{\bf{X}}^T}{\bf{A}} = {\bf{I}}}
	\end{array}$
	
	\KwIn{${{\bf{X}}_{\bf{s}}},\;\;{{\bf{X}}_{\bf{t}}},\;\;{{\bf{M}}_{{\bf{RSA}}}},\;\;{\lambda _{1\;}}\;and\;\;{\lambda _2}$}
	\textbf{Initialization}:\resizebox{0.85\hsize}{!}{%
		${{\bf{Z}}_{\bf{s}}} = {{\bf{Z}}_{\bf{l}}} = {\bf{Z}} = {\bf{0}};\;{\bf{E}} = {\bf{0}};\;{{\bf{Y}}_{\bf{1}}} = {{\bf{Y}}_{\bf{2}}} = {{\bf{Y}}_{\bf{3}}} = {\bf{0}};\;\mu  = 0.18;\;\;\varepsilon  = {10^{ - 7}};\;{\mu _{\max }} = {10^8};\rho  = 1.01$}

	\While{$\sim isempty(\bf{X},{{\bf{Y}}_{\cal S}})$ and not converged	}{
		\textbf{Step {1}}:  Fix all values except $A$ and update $A$ by setting
		  		\resizebox{0.85\hsize}{!}{%
		  			${{\bf{A}}^{{\rm{new}}}} = {(2{\bf{X{{\bf{M}}_{{\bf{RSA}}}}}}{{\bf{X}}^T} + 2\lambda {\bf{I}} + \mu ({{\bf{X}}_t} - {{\bf{X}}_s}{\bf{Z}}){({{\bf{X}}_t} - {{\bf{X}}_s}{\bf{Z}})^T} + \mu {\bf{XH}}{{\bf{X}}^T})^{ - 1}}\mu ({{\bf{X}}_t} - {{\bf{X}}_s}{\bf{Z}}){({\bf{E}} - \frac{{{{\bf{Y}}_1}}}{\mu })^T})$}
		  			\\
				  		\vspace{1pt}
		\textbf{Step 2}: Fix all values except ${{\bf{M}}_{{\bf{RSA}}}}$ and update ${{\bf{M}}_{{\bf{RSA}}}}$ by setting \\
		  		(i)Embed data via the transformation, ${{\bf{D}}_{{\bf{mmd}}}} = {{\bf{A}}^T}{\bf{X}}$. \\
		  		(ii) train a classifier $f$ and update pseudo target labels ${\bf{Y}}_{\cal T}$\\
		  		 (iii) update ${\bf{M}}_{\bf{C}}$ and ${{{\bf{M}}_{\bf{REP}}}}$ via Eq.(\ref{eq:Mmatrix}) and Eq.(\ref{eq:MC});\\
		  		(iv) obtain ${\bf{M}}_{\bf{RSA}}^{new}$\\
	\textbf{Step 3}: Fix all values except ${\bf{Z}}$ and update ${\bf{Z}}$ by setting \\
			${{\bf{Z}}^{new}} = (({{\bf{Z}}_{\bf{l}}} + {{\bf{Z}}_{\bf{s}}} - \frac{{{{\bf{Y}}_2} + {{\bf{Y}}_3}}}{\mu }) - {{\bf{X}}_s}^T{\bf{A}}({{\bf{A}}^T}{{\bf{X}}_t} - {\bf{E}} + \frac{{{{\bf{Y}}_1}}}{\mu })){(\mu {{\bf{X}}_s}^T{\bf{A}}{{\bf{A}}^T}{{\bf{X}}_s} + 2\mu {\bf{I}})^{ - 1}}$
			\vspace{2pt}
			
		\textbf{Step 4}: Fix all values except ${\bf{Z}}_{\bf{l}}$ and update ${\bf{Z}}_{\bf{l}}$ by setting\cite{DBLP:journals/pami/LiuLYSYM13,DBLP:journals/tip/XuFWLZ16} \\
    		\resizebox{1\hsize}{!}{%
    			$\begin{array}{l}
    			{\bf{Z}}_{\bf{l}}^{{\bf{new}}} = {\vartheta _{1/\mu }}({\bf{Z}} + \frac{{{{\bf{Y}}_2}}}{\mu })\;\;
    			s.t.\;{\vartheta _\lambda }({\bf{X}}) = {\bf{U}}{{\bf{S}}_\lambda }(\sum ){{\bf{V}}^T},\;{{\bf{S}}_\lambda }({\sum _{ij}}) = {\mathop{\rm si}\nolimits} gn({\sum _{ij}})\max (0,\left| {{\sum _{ij}} - \lambda } \right|),\;{\bf{X}} = {\bf{U}}\sum {{\bf{V}}^T}
    			\end{array}$}
	\vspace{2pt}
	\textbf{Step 5}: Fix all values except ${\bf{Z}}_{\bf{s}}$ and update ${\bf{Z}}_{\bf{s}}$ by setting\cite{DBLP:journals/pami/LiuLYSYM13} \\
    $\begin{array}{*{20}{c}}
    {{\bf{Z}}_{\bf{s}}^{new} = sign\;\max ((\left| {{\bf{Z}} + \frac{{{{\bf{Y}}_3}}}{\mu }} \right| - \frac{{{\lambda _2}}}{\mu }),0)}
    \end{array}$
	\vspace{2pt}
	
 	\textbf{Step 6}: Fix all values except $\bf{E}$ and update $\bf{E}$   by setting\cite{DBLP:journals/pami/LiuLYSYM13} \\   	
    $\begin{array}{c}
    	  			{{\bf{E}}^{new}} = sign\;\max ((\left| {{{\bf{A}}^T}{{\bf{X}}_{\bf{t}}} - {{\bf{A}}^T}{{\bf{X}}_s}{\bf{Z}} + \frac{{{{\bf{Y}}_1}}}{\mu }} \right| - \frac{{{\lambda _1}}}{\mu }),0)
    	  		\end{array}$
    	  		
 	\textbf{Step 7}: Update the multipliers and parameter by\cite{DBLP:journals/ijcv/ShaoKF14,DBLP:journals/tip/XuFWLZ16}\\   	
    	$\left\{ {\begin{array}{*{20}{l}}
    		{{{\bf{Y}}_1}{\rm{ = }}{{\bf{Y}}_1}{\rm{ + }}\mu ({{\bf{A}}^T}{{\bf{X}}_{\bf{t}}} - {{\bf{A}}^T}{{\bf{X}}_s}{\bf{Z}} - {\bf{E}});\;{{\bf{Y}}_2}{\rm{ = }}{{\bf{Y}}_2}{\rm{ + }}\mu \;({\bf{Z}} - {{\bf{Z}}_{\bf{l}}})}\\
    		{{{\bf{Y}}_3}{\rm{ = }}{{\bf{Y}}_3}{\rm{ + }}\mu \;({\bf{Z}} - {{\bf{Z}}_{\bf{s}}});\;\;\mu  = \min (\rho \mu ,{\mu _{\max }})}
    		\end{array}} \right.$
	\vspace{2pt}
	
 	\textbf{Step 8}: Check the convergence conditions\\   	
 	${\left\| {{{\bf{A}}^T}{{\bf{X}}_{\bf{t}}} - {{\bf{A}}^T}{{\bf{X}}_s}{\bf{Z}} - {\bf{E}}} \right\|_\infty } < \varepsilon ;\;{\left\| {{\bf{Z}} - {{\bf{Z}}_{\bf{l}}}} \right\|_\infty } < \varepsilon ;\;{\left\| {{\bf{Z}} - {{\bf{Z}}_{\bf{s}}}} \right\|_\infty } < \varepsilon \;$
}
\KwOut{Adaptation matrix ${{\bf{A}}_{{\bf{alm}}}} = {\bf{A}}$, embedding ${\bf{D}}_{{\bf{alm}}}^{{\bf{new}}} = {\bf{A}}_{{\bf{alm}}}^{\bf{T}}{\bf{X}}$}
\end{algorithm}

\section{Experiments}
In this section, we verify the effectiveness of our proposed  domain adaptation model, \textit{i.e.}, RSA-CDDA, on 16 cross-domain image classification tasks.

\subsection{Benchmarks,  Baseline Methods and Experimental setup}
In DA, USPS+MINIST, COIL20E and office+Caltech are standard benchmarks for the purpose of evaluation and comparison with state of the art. In this paper, we follow the data preparation as most previous works. We construct 16 datasets for different image classification tasks. They are: (1) the \textbf{USPS} and \textbf{MINIST} datasets of digits, but with different data distributions. We build the cross-domains as: \emph{USPS vs MNIST} and \emph{MNIST vs USPS}; (2) the \textbf{COIL20} dataset with 20 classes, split into \emph{ COIL1 vs COIL2} and \emph{COIL2 vs COIL1}; (3) \textbf{Office} and \textbf{Caltech-256}. Office contains three real-world datasets: \textbf{Amazon}(images downloaded from online merchants), \textbf{Webcam}(low resolution images) and \textbf{DSLR}( high-resolution images by digital web camera). \textbf{Caltech-256} is a standard dataset for object recognition, which contains 30,607 images for 31 categories. We denote the dataset \textbf{Amazon},\textbf{Webcam},\textbf{DSLR},and \textbf{Caltech-256} as \textbf{A},\textbf{W},\textbf{D},and \textbf{C}, respectively.  $4\times 3=12$ domain adaptation tasks can then be constructed, namely \emph{A} $\rightarrow$ \emph{W} $\dots$ \emph{C} $\rightarrow$ \emph{D}, respectively.

The proposed RSA-CDDA  is compared with nine methods of the literature, excluding only CNN-based works, given the fact that we are not using deep features. They are: (1)1-Nearest Neighbor Classifier(NN); (2) Principal Component Analysis (PCA) +NN; (3) Geodesic Flow Kernel(GFK) \cite{gong2012geodesic} + NN; (4) Transfer Component Analysis(TCA) \cite{pan2011domain} +NN; (5)Transfer Subspace Learning(TSL) \cite{4967588} +NN; (6) Joint Domain Adaptation (JDA) \cite{long2013transfer} +NN. (7) Close yet discriminative domain adaptation (CDDA)\cite{DBLP:journals/corr/LuoWHWTC17}  +NN. (8) Low-rank and Sparse Representation (LRSR) \cite{DBLP:journals/tip/XuFWLZ16} +NN. (9) Low-rank Transfer subspace Learning (LTSL)\cite{DBLP:journals/ijcv/ShaoKF14} +NN.  Note that TCA and TSL can be viewed as special case of JDA with $C=0$, and JDA a special case of CDDA method when the \textit{repulsive force} domain adaptation is ignored.  

All the reported performance scores of the eight methods of the literature are directly collected from the authors' publication. Please note that partial experimental results are quoted from CDDA and LRSR. They are assumed to be their \emph{best} performance. 


In terms of experimental setup, it is not possible to tune the set of optimal hyper-parameters, given the fact that the target domain has no labeled data. Following the setting of CDDA, LRSR and LTSL, we also evaluate the proposed RSA-CDDA by empirically searching the parameter space for the \emph{optimal} settings. Specifically, the proposed \textbf{Algorithm 1(a)}  has two hyper-parameters, \textit{i.e.}, the subspace dimension $k$, regularization parameters $\lambda $ . In  our experiments, we set $k = 100$ and 1) $\lambda  = 0.1$ for \textbf{USPS}, \textbf{MNIST} and \textbf{COIL20} , 2) $\lambda  = 1$ for \textbf{Office} and \textbf{Caltech-256}. In \textbf{Algorithm 1(b)} there are three hyper-parameters, \textit{i.e.}, the subspace dimension $k$, regularization parameters ${\lambda _1}$ and ${\lambda _2}$. In  our experiments, we set $k = 10$ and the remaining parameters similar to those in LRSR.

In our experiment, {\emph{accuracy}}  on the test dataset is the evaluation measurement. The accuracy definition is $Accuracy = \frac{{\left| {x:x \in {D_T} \wedge \hat y(x) = y(x)} \right|}}{{\left| {x:x \in {D_T}} \right|}}$, which is widely used in the literature, \textit{e.g.},\cite{pan2008transfer,long2013transfer,long2015learning}, \textit{etc}. ${\cal{D_T}}$ is the target domain treated as test data, ${\hat{y}(x)}$ is the predicted label and ${y(x)}$ is the ground truth label for a test data  $x$.

\subsection{Experimental Results and Discussion}
\begin{table}[h!]
	\centering
	\label{tab:acc}
	\caption{Quantitative comparisons with the baseline methods: Accuracy($\% $) on 16 cross-domain image classifications on seven different datasets}
	\vspace{1pt}
	\resizebox{1\columnwidth}{!}{%
		\begin{tabular}{|l |c |c |c| c |c |c | c| c| c| c|}\hline
			\toprule
			
			Datasets& {NN} & {PCA} & {GFK} & {TCA} & {TSL}  & {JDA} & \textbf{CDDA} & \textbf{LRSR} & \textbf{LTSL} & \textbf{RSA-CDDA} \\
			\midrule
			
			USPS \emph{vs} MNIST& 44.70&	44.95&	46.45&	51.05& 53.75&	59.65&	62.05& 52.33	&	&	\textbf{ 63.20}
			\\
			MNIST \emph{vs} USPS& 65.94&	66.22&	67.22&	56.28&	66.06&	67.28&	76.22& 58.55&	&	\textbf{77.50}
			\\
			\hline
			COIL1 \emph{vs} COIL2& 83.61&	84.72&	72.50&	88.47&	88.06&	89.31&	91.53&	88.61&	 75.69&	\textbf{ 95.42}
			\\
			COIL2 \emph{vs} COIL1& 82.78&	84.03&	74.17&	85.83&	87.92&	88.47&	93.89&	 89.17&	 72.22&	\textbf{ 95.28}
			\\
			\hline

			C $\rightarrow$ A&  23.70&	36.95&	41.02&	38.20&	44.47&	44.78&	48.33&	\textbf{ 51.25}&	25.26&	 45.30
			\\
			C $\rightarrow$ W&  25.76&	32.54&	40.68&	38.64&	34.24&	41.69&	\textbf{44.75}&	38.64&	 19.32&	 41.69
			\\
			C $\rightarrow$ D&  25.48&	38.22&	38.85&	41.40&	43.31&	45.22&	48.41&	47.13&	21.02&	\textbf{ 49.04}
			\\
			A $\rightarrow$ C&  26.00&	34.73&	40.25&	37.76&	37.58&	39.36&	42.12&	\textbf{ 43.37}&16.92&	39.09
			\\
			A $\rightarrow$ W&  29.83&	35.59&	38.98&	37.63&	33.90&	37.97&41.69&	36.61&	 14.58&	\textbf{ 43.39}
			\\
			A $\rightarrow$ D&  25.48&	27.39&	36.31&	33.12&	26.11&	\textbf{39.49}&	37.58&	38.85&	 21.02&	\textbf{ 39.49}
			\\
			W $\rightarrow$ C&  19.86&	26.36&	30.72&	29.30&	29.83&	31.17&	31.97&	 29.83&	 \textbf{34.64} &	 32.95
			\\
			W $\rightarrow$ A&  22.96&	31.00&	29.75&	30.06&	30.27&	32.78&	37.27&	 34.13&	\textbf{ 39.56}&	 35.28
			\\
			W $\rightarrow$ D&  59.24&	77.07&	80.89&	87.26&	87.26&	89.17&	87.90& 82.80&	72.61&	\textbf{ 94.90}
			\\
			D $\rightarrow$ C&  26.27&	29.65&	30.28&	31.70&	28.50&	31.52&	 34.64&	31.61&	\textbf{ 35.08}&	33.66
			\\
			D $\rightarrow$ A & 28.50&	32.05&	32.05&	32.15&	27.56&	33.09&	33.51& 33.19&	\textbf{ 39.67}&	36.01
			\\
			D $\rightarrow$ W & 63.39&	75.93&	75.59&	86.10&	85.42&	89.49&	\textbf{90.51}&	 77.29&	 74.92&	90.17
			\\
			\hline
			\hline
			Average (USPS)& 55.32&	55.59&	56.84&	53.67&	59.90&	63.47&	69.14&	55.44 &	\textbf{ }&	\textbf{ 70.35}
			\\
			Average (COIL)& 83.20&	84.38&	73.34&	87.15&	87.99&	88.89&	92.71&	 88.89&	 73.96&	\textbf{ 95.35}
			\\
			Average (Amazon)&  31.37&	39.79&	42.95&	43.61&	42.37&	46.31&	48.22&	45.39&	 34.55&	\textbf{48.41}
			\\
			
			Overall Average&     40.84 &	47.34&	48.48 &	50.31&	50.27&	53.78 &	56.40&	52.09 &	\textbf{ }&	\textbf{ 57.02}
			\\
			\bottomrule
		\end{tabular}}
		
	\end{table}

      The classification accuracies of the proposed method and the nine baseline methods are shown in Table.1. The highest accuracy for each cross-domain adaptation task is highlighted in bold. For fair comparison, all methods listed in Table.1 are proposed to use nearest neighbor classifier. As shown in Table.1, the proposed method depicts an overall average accuracy of $57.02\% $, which outperforms the nine baseline algorithms. The proposed method ranks first in terms of accuracy on 9 cross-domain tasks out of 16, and achieves the best average accuracy on the three datasets as well as the best overall average, thereby demonstrating the effectiveness of the proposed method. It is worth noting that the proposed method depicts $95.35\% $ accuracy on COIL20. This is rather an unexpected impressive score given the unsupervised nature of the domain adaptation for the target domain.


\section{Conclusion and Future Work}
In this paper, we have proposed a novel DA method, namely RSA-CDDA, which brings closer both marginal and class conditional data distributions between source and target and  aligns inherent hidden source and target data geometric structures while achieving discriminative DA in repulsing interclass source and target data. Comprehensive experiments on 16 cross-domain datasets for image classification task verify the effectiveness of the proposed method in comparison with nine baseline methods of the literature.  Our future work will concentrate on embedding the proposed method in  deep  networks and study other vision tasks, \textit{e.g.}, object detection, within the setting of transfer learning.

\small\bibliographystyle{plain}
\small\bibliography{egbib}

\end{document}